# A dissimilarity-based approach to predictive maintenance with application to HVAC systems

Riccardo Satta, Stefano Cavallari, Eraldo Pomponi,
Daniele Grasselli, Davide Picheo, Carlo Annis
{rsatta, scavallari, epomponi}@cgnal.com
{daniele.grasselli, davide.picheo, carlo.annis}@myefm.it

*Abstract*
*The goal of predictive maintenance is to forecast the occurrence of faults of an appliance, in order to proactively take the necessary actions to ensure its availability. In many application scenarios, predictive maintenance is applied to a set of homogeneous appliances. In this paper, we firstly review taxonomies and main methodologies currently used for condition-based maintenance; secondly, we argue that the mutual dissimilarities of the behaviours of all appliances of this set (the "cohort") can be exploited to detect upcoming faults. Specifically, inspired by dissimilarity-based representations, we propose a novel machine learning approach based on the analysis of concurrent mutual differences of the measurements coming from the cohort. We evaluate our method over one year of historical data from a cohort of 17 HVAC (Heating, Ventilation and Air Conditioning) systems installed in an Italian hospital. We show that certain kinds of faults can be foreseen with an accuracy, measured in terms of area under the ROC curve, as high as 0.96.*

## 1. Introduction

Predictive maintenance refers to the ability to forecast imminent breakdowns and faults of an appliance, in order to proactively take the necessary actions to ensure its availability and smooth operation. A general approach to the problem is to model the behaviour of the monitored appliance in terms of inputs, outputs and any kind of telemetry/sensor data that may be available. Such a model is then used to recognise anomalous behaviours and the imminence of failure.

In many application scenarios, predictive maintenance is used on a set of homogeneous appliances (e.g., all air conditioning systems in a building). We call this set the "cohort" of appliances. In this paper, we argue that the mutual dissimilarities between appliances' behaviours can reveal an upcoming fault with enough anticipation to allow for a proactive intervention and avoid interruption of operation. Specifically, we propose a novel machine learning approach based on the analysis of mutual differences of the measurements coming from the cohort. We drew inspiration from an existing parading in the field of pattern recognition, more specifically dissimilarity-based representations [1]. The model we devised can learn how these differences relate to faults by training on historical data.

The presented method has an important advantage with respect to the standard machine learning approach. Indeed, our technique is agnostic with respect to the physical meaning and behaviour of sensor inputs, whereas more standard approaches typically imply handcrafting opportune features based on the raw sensor input, in order to help a classifier to recognise the event of interest. Furthermore, by considering mutual differences instead of absolute values, the proposed approach well copes with seasonal trends and biases.

We present a preliminary evaluation of the effectiveness of this approach, conducted over one year of historical data from a cohort of 17 HVAC (Heating, Ventilation and Air Conditioning) systems installed in an Italian hospital. We train our model to early detect 4 different types of faults. We show that certain kinds of faults can be foreseen with a one-week advance with a True Positive Rate as high as almost 100% and a corresponding False Positive Rate lower than 5%. Tests on other faults show a lower but nevertheless valuable accuracy, especially in cases where a certain amount of false positives can be tolerated (e.g., when the cost of a maintenance visit is risible).

The remainder of the paper is structured as follows. We review the scientific and technical literature on the topic in [Section 2](). In [Section 3]() we present our dissimilarity-based cohort analysis framework for predictive maintenance. We then describe how we applied the proposed framework and show the obtained prediction accuracy in [Section 4](). Finally, in [Section 5]() we sum up our work and outline future research directions.

## 2. Previous Work

Eisenmann and Eisenmann [2] proposed a taxonomy of maintenance approaches consisting of three main categories: 1) *corrective maintenance* or *reactive maintenance*, i.e., repair after a failure has occurred; 2) *preventive maintenance*, i.e., take preventive actions such as periodic inspections and replacements; 3) *predictive maintenance* or *condition-based maintenance* (CBM), i.e., attempt to forecast the imminent occurrence of a failure and intervene before it happens.

Predictive maintenance itself can be declined in two main approaches [3,4]:
   a) **Phenomenological (or data-driven) approach**: the detection of an imminent failure is performed by learning the difference between normal and anomalous behaviours from the available historical data.
   b) **Model-based approach**: a model of the appliance is created (e.g. based on its original blueprints and schematics), then an imminent failure is detected by comparing the actual behaviour with the simulated one.

Model-based approaches exploit more knowledge of the monitored process than phenomenological ones, and should therefore exhibit a higher accuracy in predicting the imminence of faults [5–11]. However, accurate modelling of complex systems such as industrial appliances is most often unpractical if not unfeasible at all.

Phenomenological approaches are easier to put in place and leverage on data analysis techniques to model systems "from the outside" exploiting domain knowledge when possible. Thanks to the massive development of machine learning in recent years, phenomenological approaches have lead to many success cases. Our proposed method falls into this category; in the remainder of this section we survey the current scientific literature on the topic.

Many research papers (e.g. [12–16]) distinguish between two kinds of condition-based maintenance, namely *diagnosis* and *prognosis*. The former consists in determining whether something is going wrong in the monitored appliance ("fault detection"), in locating the faulty component ("fault isolation") and/or determine the nature of the fault ("fault identification"). The latter specialises on forecasting the occurrence of a fault. More precisely, as defined by the International Organization for Standardization [17], "prognostics is the estimation of time to failure and risk for one or more existing and future failure modes", i.e., estimation of the time remaining before a failure (Remaining Useful Life – RUL) and/or of the probability that a systems operates without failing until the next inspection (or predicted monitoring) interval [14].

Prognostics has a higher economical value as its goal is to achieve zero-downtime performance. Nevertheless diagnostics is important when prognostics fails and a fault occurs [14]. Hereafter we focus on *prognostics-aimed predictive maintenance*.

While some effort has been spent in defining common frameworks for predictive maintenance (e.g., see [13,18,19]), the vast majority of research literature targets specific maintenance problems and appliance categories. Still, we can outline the following basic steps:

- **Identification of input variables**. All sensors embedded into the appliance (e.g. temperature, pressure), as well as any measurable input or output (e.g. power consumption).
- **Identification of target(s)**. A predictive maintenance system may be aimed at forecasting the occurrence of a specific type of failure within a certain time span, the Remaining Useful Life (RUL) – Time to Failure (TTF) or similar targets.
- **Modelling of the relation between target(s) and inputs**. An appropriate model is built, optimised and validated using historical data.
- **Decision making**. A process must be put in place to properly react to predictions made by the model, e.g. establishing how to avoid the predicted failure, when to schedule the repair, or how to reschedule preventive maintenance.

Jardine et al. [14] provide good information on these steps, in particular on the possible kinds of input variables (events, measurements), on their fusion (data-level, feature-level, decision-level) and on the general techniques for processing them and building models.

The last step, *decision making*, is possibly the most crucial one, as it realises the integration between predictive maintenance and the decisional process. Criteria such as risk, cost, reliability and availability must be considered in order to deliver true value to the industry; indeed predictive maintenance is a non-trivial task both from a business and from a data science perspective.

Regarding the *business perspective*, setting up a predictive maintenance system clearly has a cost in terms of time and money invested. Unfortunately, it is still difficult to estimate *ex-ante* the Return of Investment (ROI) compared to standard maintenance approaches (corrective and preventive). Some efforts have been made in this direction targeting specific scenarios (see e.g. [20]), however a general framework is still missing. A way to assess the economic viability of predictive maintenance relies on estimating the cost of the errors made by the model. In [21] the authors consider the two possible kinds of errors related to fault forecasting:

- *Unnecessary maintenance (UM)*, or *Type I Error*, or *False Positive*. A faulty condition is erroneously predicted, resulting in an unnecessary maintenance intervention.
- *Un-prevented out-of-control state (UOC)*, or *Type II Error*, or *False Negative*. A fault occurs without being forecasted/detected, thus requiring corrective maintenance.

The two errors *UM* and *UOC* have associated unit-costs. If we denote them respectively as $C_{UM}$ and $C_{UOC}$, and with $n_{UM}$ and $n_{UOC}$ their expected number over a period of time, the total cost is easily computed as:

$$C = n_{UM} \times C_{UM} + n_{UOC} \times C_{UOC}$$

This model allows for a rough comparison between a predictive maintenance approach (where $n_{UM}$ and $n_{UOC}$ can be estimated from a test set of historical data) and classical corrective maintenance (where $n_{UM} = 0$ and therefore the total cost is entirely proportional to $n_{UOC} = n_{FAULT}$). The model can be also used to compare different predictive maintenance methods, and/or to choose the optimal hyper-parameters, *w*, of a machine learning model (in this case $n_{UM}$ and $n_{UOC}$ become a function of *w*). Lastly, if the machine learning model output is a probability of fault rather than a sharp, binary decision, the cost model can be conveniently used to estimate the decision threshold, *Th*, (i.e., the probability value above which an intervention is scheduled); in this case, $n_{UM}$ and $n_{UOC}$ become a function of *Th*. The same considerations hold for the cost model used in [22,23], where the two error indicators are Unexpected Breaks and Unexploited Lifetime.

Regarding *data science* aspects, three challenges are worth a particular mention:

- *Scarcity of training data*. Historical data is used in data-driven approaches to learn how to recognise a fault. Unfortunately, the amount of such data is often scarce [24], especially with respect to the number of positive samples (i.e. faults of the analysed appliance). A related problem is that many existing training datasets are generated ad-hoc in *simulated conditions* rather than acquired during real operation in industrial sites.
- *High dimensional data*. Industrial appliances may have many sensors, inputs and outputs. It is generally difficult to say a-priori

- which of these variables are relevant for fault forecasting. Domain knowledge can help in certain cases; otherwise, to avoid over-fitting problems caused by high dimensional data coupled with data scarcity, some statistical (e.g. correlation, Anova) and dimensionality reduction techniques (e.g. PCA, LDA) can be used. An alternative way to face this issue is to adopt dissimilarity-based representations as proposed in this work.
- *Little or no information on internal state* (i.e. only indirect information available). The internal state of an appliance can be very important in understanding the dynamics of failures. However, in many cases industrial appliances can only be treated as black boxes, especially if already installed and in operation.

We focus on the data-driven approach, which, as stated above, attempts to derive models from collected sensor and event data [14]. In this field there is a great variety of possible approaches in terms of feature engineering and classification algorithms used.

Many authors use *regression techniques* to estimate the remaining useful life time before a failure (prognosis). E.g., in [24] regression is used to this purpose on a Ion Beam Etching process for semiconductor manufacturing. Regression is often used as a final step on a multistage approach. In [12], a decision tree classifier is built based on 1000 instantaneous features (with a data selection step to reduce their number to 100) to predict three possible classes (no maintenance needed, corrective maintenance needed, preventive maintenance needed). In the case the latter class is predicted, regression is used to estimate the remaining time to failure. The technique is applied on copy machines and high-end microscopes. A two-stage approach is also used in [16], where at first samples are labelled as *normal* or *potential incipient failure* using a binary classifier. The second stage is run only for the latter class, and consists in an N-class classification corresponding to N pre-defined time windows in the future where the fault is expected to happen. Regression is performed optionally to more closely estimate the time to failure within the predicted time window. In [25], an unsupervised clustering approach is used to detect early degradation of a vertical form seal and fill (VFFS) machine. A nonlinear auto-regressive model is then built to predict some indicators of machine degradation.

The remaining useful life and other maintenance indicators can be also estimated using a *neural network* (NN), e.g., in [26], where NNs are exploited to estimate the condition of a cutting tool from indirect measurements based on cutting force. Statistical techniques (ANOVA and correlation coefficient) are used to select the most effective features and training samples. In [27] dynamic wavelet neural networks are applied to estimate the remaining useful life as the time left before the fault. In [28] vibration signals are used to estimate the failure time of bearings. A two step approach is adopted: first, several neural networks are trained, each on different bearings in a training set; then, NN outputs are combined linearly using weights learned on a validation set of bearings to obtain the final failure time estimation. They used three different weighting algorithms and two types of neural network models (single-bearing and clustered-bearing). In [29], NNs are used to estimate both the life percentile and failure times. In particular, a cost matrix and a probabilistic replacement model that optimises the expected cost per unit time is developed. In [30] a mathematical framework is developed to estimate the optimal monitoring intervals for 2-phase systems. A 2-phase system initially operates in a new condition for a time span T, before evolving to a worn condition where it resides for time span W preceding system failure. The expected cost is decomposed into 2 components: one due to maintenance actions, the other due to monitoring actions. Then, by assuming the distribution for T and W for both fixed and non-fixed monitoring-interval policies, they derive the intervals that minimise the total cost-rate.

Another class of techniques tries to predict classes rather than indicators, using standard classifiers such as SVM and Decision Trees. In [31], a novelty detection technique based on One-Class SVM is developed to early detect 4 types of faults of a HVAC chilling system. Input features for the classifier are obtained using a steady-state data filter applied to three sensor signals. Also handcrafted features are used. In [9] a diagnosis system based on boosted decision trees is used to predict the occurrence of a failure of the air pressure system of an heavy truck. Various statistical techniques are used to generate features from 170 continuous variables (e.g., box-plot analysis, outlier detection). In [23], an ensemble of classifiers (either SVM or kNN) with different prediction horizons are trained, to provide a choice of trade-offs in terms of frequency of unexpected breaks. Statistical features (min, max, average, etc.) are used to describe time-series of data coming from different sensors.

### 3. DISSIMILARITY-BASED COHORT ANALYSIS FOR PREDICTIVE MAINTENANCE

We present here our approach to predictive maintenance using dissimilarity-based representations and cohort analysis. Our approach is intended for a scenario in which predictive maintenance is applied to a set of homogeneous[1] appliances, in operation simultaneously, all monitored by a set of sensors measuring the same quantities (e.g., temperature, pressure, power consumption, etc.). Examples of such sets are the elevators or the air conditioning systems installed in a large building. We refer to this homogeneous appliances as the "cohort" (of elevators, of air conditioning systems, etc.).

In this work, we argue that the *concurrent mutual differences* among the elements of the cohort encode potentially useful information for predictive maintenance. To the best of our knowledge all previous studies consider instead only one appliance, or each appliance separately from any other. We are driven in this assumption by the body of work on dissimilarity representations [1], a kind of representation where objects are represented by their mutual dissimilarity[2]. Such representations have been proven successful especially in classification problems where it is difficult to define meaningful features and/or a simple statistical model, such as the case at hand [12,19].

---

[1] With the term homogeneous here we intend mechanically identical appliances, e.g. of the same brand and model.
[2] The way dissimilarity is quantified depends case by case on the model and the objects treated.

In addition, we hypothesise that anomalies are more easily detected by looking at dissimilarities among homogeneous appliances than when considering each appliance in isolation. See e.g. the toy example in Figure 1 and Figure 2, which depicts the behaviour of four homogeneous appliances in the same "sensor space". During normal operation (Figure 1), it is likely to find similar behaviours in all four appliances; consequently, the mutual dissimilarities among them should remain more or less constant. When one appliance shows an anomalous behaviour (Figure 2), this is reflected by a change in the mutual dissimilarities. A classifier can then learn which patterns of changes in mutual dissimilarities can lead to a fault.

It is worth to point out two important advantages of our method with respect to standard machine learning approaches. First, dissimilarity-based representations are agnostic to the physical meaning of sensor inputs, and therefore does not rely on handcrafting opportune features based on the raw sensor inputs as many other methods do. Second, by considering mutual differences in place of absolute values, our technique is robust to seasonal trends and biases due to exogenous factors.

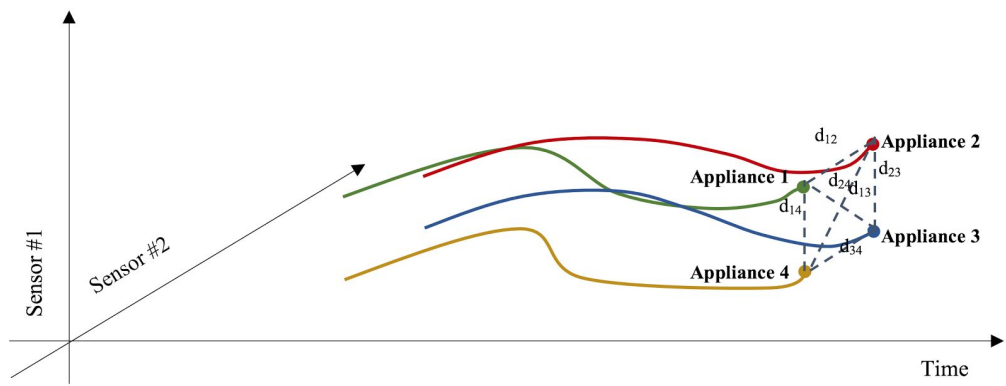

Figure 1: Toy example. The four coloured curves represent the behaviour over time of four appliances with respect to two sensors. In normal operation, one expects the four behaviours to be similar (if not equal), therefore the relative distances should show modest variations over time.

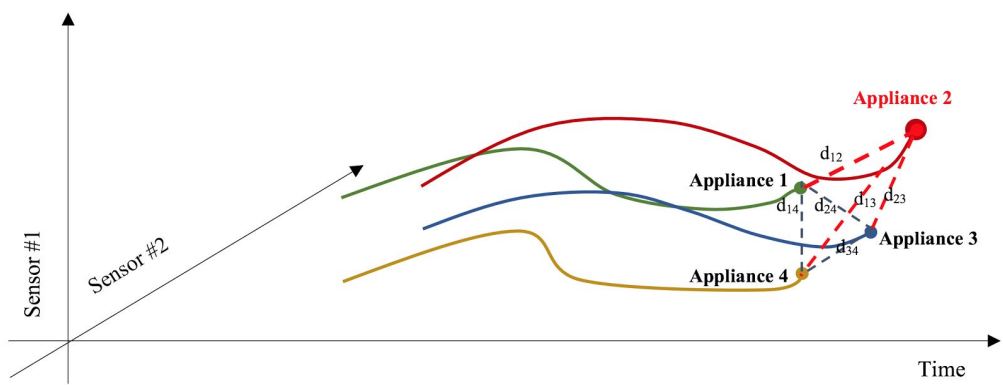

Figure 2: Toy example (cont.). Appliance 2 shows an anomalous behaviour, which is likely to be connected with an imminent fault. This is clearly reflected by a change in the relative distances.

Let us now develop the idea of dissimilarity representation of cohorts of appliances in a more formal way. First, we define A = {$a_1$, …, $a_N$} as the set of *N* homogeneous appliances, i.e., the cohort. Each appliance $a_i$ has *M* associated sensors $s_i^1$, …, $s_i^M$, each recording a physical measure (the resulting signals are the *telemetry* of $a_i$ ).

Inspired by dissimilarity-based representations, at a certain time *t* each appliance can be represented as a vector $\overline{a}_i(t)$ of dissimilarities between corresponding sensors of the different appliances, $d_{ij}^k(t)$, with *k* = 1, …, *M* indicating the sensor and *i,j = 1, …, N* the appliances. Omitting the *i* subscript for the sake of simplicity, we can write:

$$\overline{a}(t) = \left[d_1^1(t), ..., d_1^M(t), d_2^1(t), ..., d_2^M(t), ..., d_N^1(t), ..., d_N^M(t)\right] \quad (1)$$

A classifier can then be trained using these dissimilarity values as input features.

Instead of computing dissimilarities at each instant *t*, it makes sense to compute dissimilarity over time windows of fixed length *T* to capture the temporal behaviour. Therefore we can write the dissimilarity term $d_{ij}^k(t)$ in the following way:

$$d_{ij}^k(t) = f([s_i^k(t-T), ..., s_i^k(t)], [s_j^k(t-T), ..., s_j^k(t)]) \quad (2)$$

where *f* is a dissimilarity measure between the two time series $[s_i^k(t-T), ..., s_i^k(t)]$ and $[s_i^k(t-T), ..., s_i^k(t)]$, e.g. a correlation coefficient.

The time window *T* sets how much "past" the classifier takes into consideration to make a prediction in *t* about a future target fault, *F* (e.g., a failure of a fan in an air conditioning system). In our framework, the temporal displacement of the forecasting depends on two other time windows $T_f$, and $T_a$, all relating to the time *t* when the prediction is generated, as depicted in Figure 3. In particular:
- $T_f$ is the forecasting window, i.e., the time window where the prediction falls;
- $T_a$ is the action window, i.e, the interval of time between *t* and $T_f$, which we named "action window" because it is in this time span that some manutention action is supposed to take place in case of a fault prediction.

Different values of *T*, $T_f$ and $T_a$ define different prediction scenarios.

In summary, given a fault *F* relative to the i-*th* appliance, the goal is to estimate at the instant *t* the likelihood that *F* is going to happen in $T_f$ by using the telemetry of *all the appliances* in the time interval [t-T, t] (e.g. the last week).

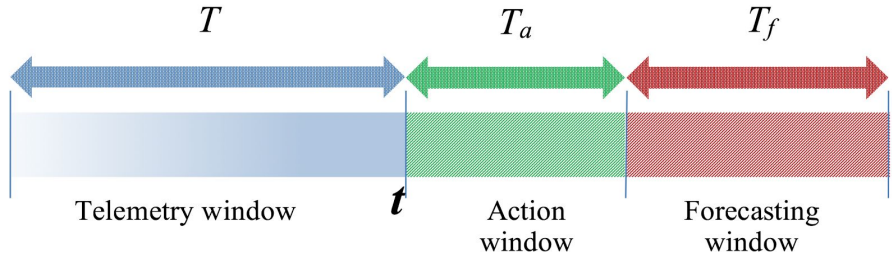

Figure 3: prediction scenario.

Note that we compute dissimilarities to signals recorded in the same instant of time (Eq. (1)) or time window (Eq. (2)). This is the reason why we say these are *concurrent* mutual differences, to stress the difference with the classical dissimilarity-based approach where differences are computed with respect to fixed prototypes [1].

4. EXPERIMENTAL EVALUATION

This section describes how we experimentally evaluated the proposed approach on a real dataset of historical data.

**Dataset.** We partnered with a leading building management company to test our approach on one year of historical data from an undisclosed Italian hospital, containing telemetry from 17 HVAC (heating, ventilation, air conditioning) appliances of surgery rooms, corridors and first-aid rooms. Each HVAC is provided with 15 sensors which monitor the tool condition by recording the signals described in Table 1.

**Preprocessing.** Each appliance can raise 22 different alarms. Out of them, we selected 4 alarms (see Table 2) according to the following two criteria: a) the technical importance from the point of view of the facility management; b) the number of positive samples in the dataset. Note that we previously applied a filter to discard false positive from the dataset, which were due to tests performed by the maintenance technical team. The sampling interval of the signals varies across the sensors, spanning from 1 min to 60 min.

We found that the sensor telemetry had some interruptions and we decided to fill the missing values with the median value of the available time-series.

As per the lengths of the telemetry, action and forecasting windows, we chose $T = 2$ weeks, $T_a = 1$ week and $T_f = 1$ week.

We created the dataset for each appliance, as follows. We slided the three windows (telemetry, action and forecasting) with steps of 1 day, and took the telemetry in $T$ at each step as one element of the dataset. We then labeled each telemetry window as positive with respect to each alarm if one such alarm fell in the forecasting window. Note that in this way a single positive alarm could appear in 7 different samples.

| Sensor Name | Monitored Variable |
|---|---|
| *ComValvFred* | Opening percentage of the electrovalve to supply the cooling battery |
| *ComValvPre* | Opening percentage of the electrovalve to supply the heating battery |
| *ComValvUmid* | Opening percentage of the electrovalve to supply the humidifier |
| *PortataMand* | Supply flow |
| *PortataRip* | Return flow |
| *PresManWilso* | Supply pressure |
| *PresRipWilso* | Return Pressure |
| *SgnIVM1* | Supply 1 fan inverter |
| *SgnIVM2* | Supply 2 fan inverter |
| *SgnIVR1* | Return 1 fan inverter |
| *SgnIVR2* | Return 2 fan inverter |
| *TempMand* | Temperature after supply fan |
| *TempRip* | Temperature before return fun |
| *UmidMand* | Humidity after supply fan |
| *UmidRip* | Humidity before return fun |

Table 1: Data recorded from the sensors mounted on each HVAC.

| Alarm | Description | # positive samples for each HVAC | # positive samples |
|---|---|---|---|
| *AlmBIVR1* | Failure of the return 1 fan inverter | (6, 0, 0, 4, 0, 0, 0, 0, 1, 1, 0, 0, 2, 7, 0, 0, 0) | 21 |
| *AlmBIVR2* | Failure of the return 2 fan inverter | (3, 0, 0, 3, 0, 0, 0, 4, 1, 0, 0, 0, 1, 7, 0, 0, 1) | 20 |
| *AlmSP* | Failure of the air shutter | (4, 4, 4, 3, 4, 0, 4, 0, 4, 3, 0, 0, 6, 0, 0, 0, 3) | 39 |
| *IntProt* | Activation of inverter thermal protection due to an increase of load current | (2, 2, 4, 3, 3, 3, 4, 1, 2, 2, 2, 2, 3, 3, 2, 3, 3) | 44 |

Table 2: Description of alarms.

**Description of the experiments.** To obtain the classification model we used the AdaBoost algorithm with decision trees as base classifiers [32]. AdaBoost is a meta-classifier which combines many weak classifiers, combined and sequentially trained in such a way that each classifier tries to improve the accuracy obtained on training samples that have been misclassified by the others. To compute the dissimilarities of Eq. (3), we used the Pearson and the Spearman correlation coefficients. We point out that such a representation of the dissimilarity is *adimensional* and completely generic, as the measure of dissimilarity does not depend on the kind of sensor. In the following, we refer to the two representations respectively as *Cohort_Pearson* and *Cohort_Spearman*. We also combined them by concatenating the two vectors *Cohort_Pearson* and *Cohort_Spearman* to form a single feature vector; this representation is referred to as *Cohort_P&S*.

For the sake of completeness, we compare our method with a baseline model built with some common (see e.g. [9,23,24,29,33]) statistical features, i.e.: max, min, average, standard deviation, skewness, kurtosis and mutual covariances of each sensor in the telemetry window. This method is referred to as *Baseline*. The *Baseline* is also combined with all the cohort based ones (this combination is referred to as *Comb*).

**Performance evaluation**. For each alarm we performed a cross-validation using as folds the historical data of each HVAC with at least one positive sample. Since the output of the model is a score associated with the telemetry window (indicating the probability of alarm), to make a decision about maintenance we set a decision threshold *Tr*, such that a score higher than *Tr* elicits the prediction of an alarm in $T_f$. Intuitively, higher values of *Tr* will result in more false negatives, whereas lower values in more false positives.

A standard way to evaluate the tradeoff between false negatives and false positives when varying the threshold is the Receiver Operating Characteristics (ROC) curve. Each point of the ROC corresponds to a value of *Tr*, and associates the attained value of Fase Positive Rate (FPR) on the *x* axis with the corresponding True Positive Rate (TPR) on the *y* axis.

We computed a ROC for each alarm and HVAC; in Figure 3 we show the average ROC over folds. Where the performance is lower (AlmBIVR1, AlmBIVR2) all the features give similar results, whereas, in the two alarms where the best performance is attained (AlmSP, IntProt), dissimilarity-based cohort features perform noticeably better than the baseline model.

Finally, in Table 3 we provide a scalar measure of performance by computing the average Area Under the ROC Curve (AUC) over the folds. Interestingly, we found that dissimilarity-based cohort approaches performed better than the baseline model in all four alarms, confirming our initial claim that anomalies can be more easily detected by evaluating mutual dissimilarities instead of considering each appliance separately. Furthermore, in three out of four cases, the best performances were obtained by the combination of all the features. This is an indication that baseline and dissimilarity-based cohort features are not completely redundant and can be combined to improve the accuracy of a model.

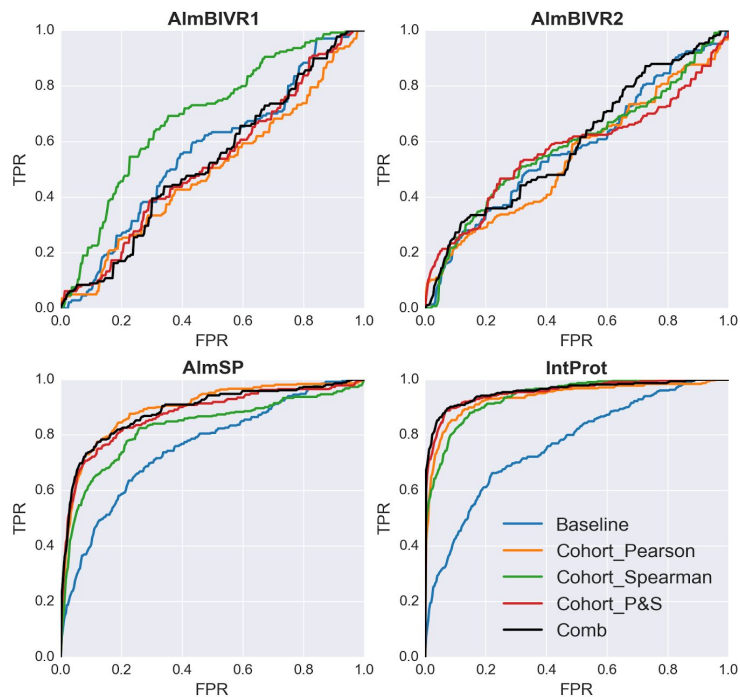

Figure 3: Each subplot shows the average ROC over folds obtained from the 5 different set of features used (see color code) and relative to the alarm specified in the title.

| Alarm | Features | Average AUC | Alarm | Features | Average AUC |
|---|---|---|---|---|---|
| *AlmBIVR1* | Baseline | 0,553 | *AlmSP* | Baseline | 0,763 |
| | Cohort_Pearson | 0,472 | | Cohort_Pearson | 0,890 |
| | Cohort_Spearman | **0,697** | | Cohort_Spearman | 0,827 |
| | Cohort_P&S | 0,537 | | Cohort_P&S | 0,874 |
| | Comb | 0,534 | | Comb | **0,899** |
| *AlmBIVR2* | Baseline | 0,561 | *IntProt* | Baseline | 0,753 |
| | Cohort_Pearson | 0,578 | | Cohort_Pearson | 0,934 |
| | Cohort_Spearman | 0,602 | | Cohort_Spearman | 0,931 |
| | Cohort_P&S | 0,626 | | Cohort_P&S | 0,950 |
| | Comb | **0,656** | | Comb | **0,956** |

Table 3: Values of the average area under the ROC curve over folds obtained for each alarm and set of features. In bold the best performance for each alarm.

## 5. Conclusions

In this paper, we presented a new method for prognostics-aimed predictive maintenance based on the dissimilarity-based representation paradigm from pattern recognition. Our method considers cohorts of identical appliances globally, rather than individually. Indeed, in each instant of time, we consider the telemetry of *all the appliances* to extract representative features of every single appliance; we claim that in this way, anomalous behaviours can be more easily detected.

We tested our dissimilarity-based features over one year of historical data from 17 Heating, Ventilation and Air Conditioning systems installed in an Italian hospital, obtaining good accuracies which improve further when combining them with standard statistical features.

Using dissimilarity-based features avoids the need to handcraft features, providing a completely agnostic (and therefore general) framework to predictive maintenance. In addition, dissimilarity features are inherently robust to biases and seasonal trends.

The attained results motivate us to further research the topic. Among the open questions, we point out three significant ones: first, it is worth to study if and how dissimilarity-based cohort features can be used to predict the Remaining Useful Life (RUL) (see Section 2); second, a complete decision making model considering also the cost of decision should be developed, similarly to [21–23]; third, other dissimilarity functions (see Eq. (2)) should be investigated.

Besides the technical future developments, it is also important to further assess the capabilities of the proposed approach with other (possibly bigger) datasets.

ACKNOWLEDGMENTS
This research has been funded by and jointly carried out with Efm srl in a proof-of-concept project on predictive maintenance.

**References**

1. Pękalska E, Duin RPW. The Dissimilarity Representation for Pattern Recognition: Foundations and Applications. World Scientific; 2005.

2. Eisenmann RC. Machinery Malfunction Diagnosis and Correction: Vibration Analysis and Troubleshooting for the Process Industries. Prentice Hall; 1998.

3. Kroll B, Bjorn K, David S, Sebastian S, Oliver N. System modeling based on machine learning for anomaly detection and predictive maintenance in industrial plants. Proceedings of the 2014 IEEE Emerging Technology and Factory Automation (ETFA). 2014. doi:10.1109/etfa.2014.7005202

4. Okoh C, Roy R, Mehnen J, Redding L. Overview of Remaining Useful Life Prediction Techniques in Through-life Engineering Services. Procedia CIRP. 2014;16: 158–163.

5. Kacprzynski GJ, Sarlashkar A, Roemer MJ, Hess A, Hardman B. Predicting remaining life by fusing the physics of failure modeling with diagnostics. JOM. 2004;56: 29–35.

6. Oppenheimer CH, Loparo KA. Physically based diagnosis and prognosis of cracked rotor shafts. Component and Systems Diagnostics, Prognostics, and Health Management II. 2002. doi:10.1117/12.475502

7. Chelidze D, David C. Multimode damage tracking and failure prognosis in electromechanical systems. Component and Systems Diagnostics, Prognostics, and Health Management II. 2002. doi:10.1117/12.475493

8. Chelidze D, David C, Cusumano JP. A Dynamical Systems Approach to Failure Prognosis. J Vib Acoust. 2004;126: 2.

9. Cerqueira V, Vítor C, Fábio P, Claudio S, Carlos S. Combining Boosted Trees with Metafeature Engineering for Predictive Maintenance. Lecture Notes in Computer Science. 2016. pp. 393–397.

10. Adams DE. Nonlinear damage models for diagnosis and prognosis in structural dynamic systems. Component and Systems Diagnostics, Prognostics, and Health Management II. 2002. doi:10.1117/12.475507

11. Ray A, Tangirala S. Stochastic modeling of fatigue crack dynamics for on-line failure prognostics. IEEE Trans Control Syst Technol. 1996;4: 443–451.

12. Abdellatif Bey-Temsamani, Marc Engels, Andy Motten, Steve Vandenplas, Agusmian P Ompusunggu. A practical approach to combine data mining and prognostics for improved predictive maintenance. Proceedings of the Third International Workshop on Data Mining Case Studies (DMCS). pp. 36–43.

13. Huang * R, Xi L, Lee J, Liu CR. The framework, impact and commercial prospects of a new predictive maintenance system: intelligent maintenance system. Prod Plan Control. 2005;16: 652–664.

14. Jardine AKS, Daming L, Dragan B. A review on machinery diagnostics and prognostics implementing condition-based maintenance. Mech Syst Signal

Process. 2006;20: 1483–1510.

15. Traore M, Chammas A, Duviella E. Supervision and prognosis architecture based on dynamical classification method for the predictive maintenance of dynamical evolving systems. Reliab Eng Syst Saf. 2015;136: 120–131.

16. Yang C, Chunsheng Y, Qiangqiang C, Yubin Y, Nan J. Developing predictive models for time to failure estimation. 2016 IEEE 20th International Conference on Computer Supported Cooperative Work in Design (CSCWD). 2016. doi:10.1109/cscwd.2016.7565977

17. ISO 13381-1, Condition monitoring and diagnostics of machines – prognostics, Part 1: General guidelines, Int. Standard, ISO. 2004.

18. Groba C, Christin G, Sebastian C, Frank R, Andreas G. Architecture of a Predictive Maintenance Framework. 6th International Conference on Computer Information Systems and Industrial Management Applications (CISIM'07). 2007. doi:10.1109/cisim.2007.14

19. Ming Tan C, Tan CM, Nagarajan R. A framework to practical predictive maintenance modeling for multi-state systems. Reliab Eng Syst Saf. 2008;93: 1138–1150.

20. Koo WL, Van Hoy T. Determining the Economic Value of Preventive Maintenance. Jones Lang LaSalle. 2003.

21. Susto GA, Alessandro B, De Luca C. A Predictive Maintenance System for Epitaxy Processes Based on Filtering and Prediction Techniques. IEEE Trans Semicond Manuf. 2012;25: 638–649.

22. Susto GA, Andrea S, Simone P, Daniele P, Sean M, Alessandro B. A predictive maintenance system for integral type faults based on support vector machines: An application to ion implantation. 2013 IEEE International Conference on Automation Science and Engineering (CASE). 2013. doi:10.1109/coase.2013.6653952

23. Susto GA, Andrea S, Simone P, Sean M, Alessandro B. Machine Learning for Predictive Maintenance: A Multiple Classifier Approach. IEEE Trans Ind Inf. 2015;11: 812–820.

24. Susto GA, Jian W, Simone P, Mattia Z, Johnston AB, O'Hara PG, et al. An adaptive machine learning decision system for flexible predictive maintenance. 2014 IEEE International Conference on Automation Science and Engineering (CASE). 2014. doi:10.1109/coase.2014.6899418

25. Langone R, Rocco L, Carlos A, De Ketelaere B, Jonas V, Wannes M, et al. LS-SVM based spectral clustering and regression for predicting maintenance of industrial machines. Eng Appl Artif Intell. 2015;37: 268–278.

26. Saglam H, Unuvar A. Tool condition monitoring in milling based on cutting forces by a neural network. Int J Prod Res. 2003;41: 1519–1532.

27. Vachtsevanos G, Wang P. Fault prognosis using dynamic wavelet neural networks. 2001 IEEE Autotestcon Proceedings IEEE Systems Readiness

Technology Conference (Cat No01CH37237). doi:10.1109/autest.2001.949467

28. Gebraeel N, Lawley M, Liu R, Parmeshwaran V. Residual Life Predictions From Vibration-Based Degradation Signals: A Neural Network Approach. IEEE Trans Ind Electron. 2004;51: 694–700.

29. Wu S-J, Sze-jung W, Nagi G, Lawley MA, Yuehwern Y. A Neural Network Integrated Decision Support System for Condition-Based Optimal Predictive Maintenance Policy. IEEE Transactions on Systems, Man, and Cybernetics - Part A: Systems and Humans. 2007;37: 226–236.

30. Maillart LM, Pollock SM. Cost-optimal condition-monitoring for predictive maintenance of 2-phase systems. IEEE Trans Reliab. 2002;51: 322–330.

31. Beghi A, Cecchinato L, Corazzol C, Rampazzo M, Simmini F, Susto GA. A One-Class SVM Based Tool for Machine Learning Novelty Detection in HVAC Chiller Systems. IFAC Proceedings Volumes. 2014;47: 1953–1958.

32. Freund Y, Schapire RE. A Short Introduction to Boosting. Journal of Japanese Society for Artificial Intelligence. 1999;14: 771–780.

33. Huda ASN, Soib T. Application of infrared thermography for predictive/preventive maintenance of thermal defect in electrical equipment. Appl Therm Eng. 2013;61: 220–227.